\documentclass[runningheads]{llncs}

\usepackage{graphicx,color}
\usepackage{xcolor}
\usepackage{calc}
\graphicspath{{figure/}}
\usepackage[ruled,linesnumbered]{algorithm2e} 
\usepackage{amsmath,bm}
\usepackage{amssymb}
\usepackage{breqn}
\usepackage{subfigure}
\usepackage{multirow}
\usepackage{psfrag}
\usepackage{url}
\usepackage{hyperref}
\usepackage{morefloats}
\usepackage{cite}
\usepackage{booktabs}
\usepackage{rotating}
\usepackage{colortbl}
\DeclareMathOperator*{\argmax}{argmax}
 
\DeclareMathOperator*{\rand}{rand} 
\usepackage{array}
\newcommand{\tabincell}[2]{\begin{tabular}{@{}#1@{}}#2\end{tabular}} 

\begin{document}

%
\title
{Defense of Word-level Adversarial Attacks via Random Substitution Encoding}
\titlerunning{Defense of Word-level Adversarial Attacks}
\author{Zhaoyang Wang \and
	Hongtao Wang\thanks{Corresponding author. Code available at: https://github.com/Raibows/RSE-Adversarial-Defense}}
\authorrunning{Z. Wang and H. Wang}
%
\institute{School of Control and Computer Engineering,\\
	North China Electric Power University, Baoding, China \\
	\email{luckychizuo@gmail.com, wanght@ncepu.edu.cn}\\
}

\maketitle

\begin{abstract}
The adversarial attacks against deep neural networks on computer vision tasks have spawned many new technologies that help protect models from avoiding false predictions.
	Recently, word-level adversarial attacks on deep models of Natural Language Processing (NLP) tasks have also demonstrated strong power, e.g.,  fooling a sentiment classification neural network to make wrong decisions.
	Unfortunately, 
	few previous literatures have discussed the defense of such word-level synonym substitution based attacks since they are hard to be perceived and detected.
	In this paper,
	we shed light on this problem and propose a novel defense framework called Random Substitution Encoding (RSE), which introduces a random substitution encoder into the training process of original neural networks.
	Extensive experiments on text classification tasks demonstrate the effectiveness of our framework on defense of word-level adversarial attacks, under various base and attack models.
\end{abstract}


\section{Introduction}
\label{sec-intro}
Deep Neural Network(DNN) has become one of the most popular frameworks to harvest knowledge from big data.
Despite their success, the robustness of DNNs has ushered a serious problem, which has prompted the adversarial attacks on them.
Adversarial attack refers to generating imperceptible perturbed examples to fool a well-trained DNN model making wrong decisions.
In the Computer Vision(CV) domain, adversarial attacks against many famous DNN models have been shown to be an indisputable threat.

Recently, adversarial attacks on DNN models for Natural Language Processing(NLP) tasks have also received significant attentions.
Existing attack methods can be classified into two categories: character-level attacks and word-level attacks.
For character-level attacks, attackers can modify several characters of an original text to manipulate the target neural network.
While character-level attacks are simple and effective, it is easy to defend when deploying a spell check and proofread algorithm before feeding the inputs into DNNs\cite{acl/PruthiDL19}.
Word-level attacks substitute a set of words in original examples by their synonyms, and thus can preserve semantic coherence to some extent. 
The adversarial examples, created by word-level attackers, are more imperceptible for humans and more difficult for DNNs to defend.

Until now, there are few works on defense of adversarial attacks against NLP tasks, e.g., text classification.
Most efforts had gone into increasing the model robustness by adding perturbations on word embeddings, e.g., adversarial training\cite{corr/GoodfellowSS14} or defensive distillation\cite{sp/PapernotM0JS16}. 
Although these approaches exhibit superior performance than base models, they assume there are no malicious attackers and could not resist word-level adversarial attacks\cite{wang2019natural}.
The only work against word-level synonym adversarial attacks is \cite{wang2019natural}.
It proposed a Synonym Encoding Method(SEM) which maps synonyms into the same word embeddings before training the deep models.
As a result, the deep models are trained only on these examples with only fixed synonym substitutions.
The reason why SEM based deep models can defend word-level attacks is that it can transform many unseen or even adversarial examples `move' towards `normal' examples that base models have seen. 
While SEM can effectively defend current best synonym adversarial attacks, it is too restrictive when the distances are large between transformed test examples and the limited training examples.

This paper takes a straightforward yet promising way towards this goal. 
Unlike modifying word embeddings before the training process, we put the synonyms substitutions into the training process in order to fabricate and feed models with more examples. 
To this end, we proposed a dynamic random synonym substitution based framework that introduces Random Substitution Encoding(RSE) between the input and the embedding layer.
We also present a Random Synonym Substitution Algorithm for the training process with RSE.
The RSE encodes input examples with randomly selected synonyms so as to make enough labeled neighborhood data to train a robust DNN. 
Note that the RSE works in both training and testing procedure, just like a dark glasses dressed on the original DNN model.

We perform extensive experiments on three benchmark datasets on text classification tasks based on three DNN base models, i.e., Word-CNN, LSTM and Bi-LSTM.
The experiment results demonstrate that the proposed RSE can effectively defend word-level synonym adversarial attacks.
The accuracy of these DNN models under RSE framework achieves better performance under popular word-level adversarial attacks, and is close to the accuracy on benign tests.


\section{Related Work}
\label{sec-relt}
Adversarial attack and defense are two active topics recently.
In natural language processing, many tasks are facing the threat of adversarial attack, e.g., Text Classification\cite{GaoLSQ18,EbrahimiRLD18-2,0002LSBLS18}, Machine Translation\cite{EbrahimiLD18}, Question \& Answer\cite{RenDW18}, etc.
Among them, text classification models are more vulnerable and become the targets of malicious adversaries.
The state-of-the-art adversarial attacks to text classification in literatures can be categorized into the following types:
\begin{itemize}
	\item Character-level attacks.
Attackers can modify a few characters of an original text to manipulate the target neural network.
Gao et al.\cite{GaoLSQ18} proposed DeepWordBug, an approach which adds small character perturbations to generate adversarial examples against DNN classifiers. 
Ebrahimi et al.\cite{EbrahimiRLD18-2} proposed an efficient method, named by Hotflip, to generate white-box adversarial texts to trick a character-level neural network.  
In \cite{0002LSBLS18}, text adversarial samples were crafted in both white-box and black-box scenarios.
However, these approaches are easy to defend by placing a word recognition model before feeding the inputs into neural network\cite{PruthiDL19}.
	\item Word-level attacks.
Word-level attacks substitute words in original texts by their synonyms so they can preserve semantic coherence. 
Liang et al.\cite{0002LSBLS18} designed three perturbation strategies to generate adversarial samples against deep text classification models.
Alzantot et al.\cite{AlzantotSEHSC18} proposed a genetic based optimization algorithm to generate semantically similar adversarial examples to fool a well-trained DNN classifier.
To decrease the computational cost of attacks, Ren et al.\cite{RenDHC19} proposed a greedy algorithm, namely PWWS, for text adversarial attack. 
Word-level adversarial examples are more imperceptible for humans and more difficult for DNNs to defend.
\end{itemize}

There exists very few works on defending word-level text adversarial attacks.
To the best of our knowledge, \cite{wang2019natural} is the only work on defenses against synonym substitution based adversarial attacks.
They proposed Synonym Encoding Method (SEM) that encodes synonyms into the same word embeddings to eliminate adversarial perturbations.
However, it needs an extra encoding stage before the normal training process and is limited on the fixed synonym substitution.
Our framework adopts a unified training process and provides a flexible synonym substitution encoding scheme.

\section{Preliminaries}
\label{sec-prel}
In this section, we firstly present the problem of adversarial attack and defense in text classification tasks.
Next we provide preliminaries about attack models: several typical word-level synonym adversarial attacks.

\subsection{Problem Definition}
\label{sec-prob}
Given a trained text classifier $F: \mathcal{X}\rightarrow\mathcal{Y}$,  $\mathcal{X}$ and $\mathcal{Y}$ denote the input and the output space respectively.
Suppose there is an input text $x\in \mathcal{X}$, the classifier can give a predicted true label $y_{true}$ based on a posterior probability $P$.
\begin{equation}\label{equ:model}
\argmax_{y_i\in\mathcal{Y}}P(y_i|x)=y_{true}
\end{equation}

An \textbf{adversarial attack} on classifier $F$ is defined that the adversary can generate an adversarial example $x'$ by adding an imperceptible perturbation $\Delta x$, such that:
\begin{equation}\label{equ:admodel}
\begin{aligned}
\argmax_{y_i\in\mathcal{Y}}P(y_i|x')&\neq y_{true} \\
s.t.\ \ x'&=x+\Delta x, \ \|\Delta x\|_p<\epsilon
\end{aligned}
\end{equation}
where $\|\cdotp\|_p$ denotes the $p$-norm and $\epsilon$ controls the small perturbation so that the crafted example is imperceptible to humans.

The \textbf{defense} against adversarial attack requires to train an enhanced text classifier $F^{\ast}$ over $F$. A successful defense means that for a given input text example $x$, the attacker failed to craft an adversarial example, or the generated adversarial example $x'$ could not fool the classifier $F^{\ast}$.
\begin{equation}\label{equ:dfmodel}
\argmax_{y_i\in\mathcal{Y}}P(y_i|x')=\argmax_{y_i\in\mathcal{Y}}P(y_i|x)=y_{true}
\end{equation}

\subsection{Synonym Adversarial Attacks}
\label{sec-atta}
To ensure the perturbation small enough, the adversarial examples need to satisfy semantic coherence constraints. 
An intuitive way to craft adversarial examples is to replace several words in the input example by their synonyms.
Let $x={w_1,\cdots,w_n}$ denote an input example, where $w_i\in W$ denotes a word.
Each word $w_i$ has a synonym candidate set $\mathcal{S}_i$.
For a synonym adversarial attack, adversary can substitute $K$ words denoted by $\mathcal{C}_x$, to craft an adversarial example $x'=w_1'\cdots w_n'$:
\begin{equation}\label{equ:saamodel}
w_i'= \left\{
\begin{aligned}
w_i &	&\text{if}\ w_i\notin \mathcal{C}_x \\
s_i^j&	&\text{if}\ w_i\in \mathcal{C}_x
\end{aligned}
\right.
\end{equation}
where $s_i^j$ denotes the $j$th substitution candidate word in $\mathcal{S}_i$.

Existing synonym substitution based adversarial attacks had gone into proposing fast searching algorithms, such as Greedy Search Algorithm(GSA)\cite{kuleshov2018adversarial} and Genetic Algorithm (GA)\cite{AlzantotSEHSC18}.
\cite{RenDHC19} proposed a fast state-of-the-art method called Probability Weighted Word Saliency(PWWS) which considers the word saliency and the classification confidence.

\section{The Proposed Framework}
\label{sec-fram}
In this section, we first present our motivation, and then demonstrate the detailed defense framework.

\subsection{Motivation}
\label{sec-moti}
\begin{figure}[tb]
	\setlength{\belowcaptionskip}{-0.5cm}
	\centering
	\includegraphics[width=\textwidth]{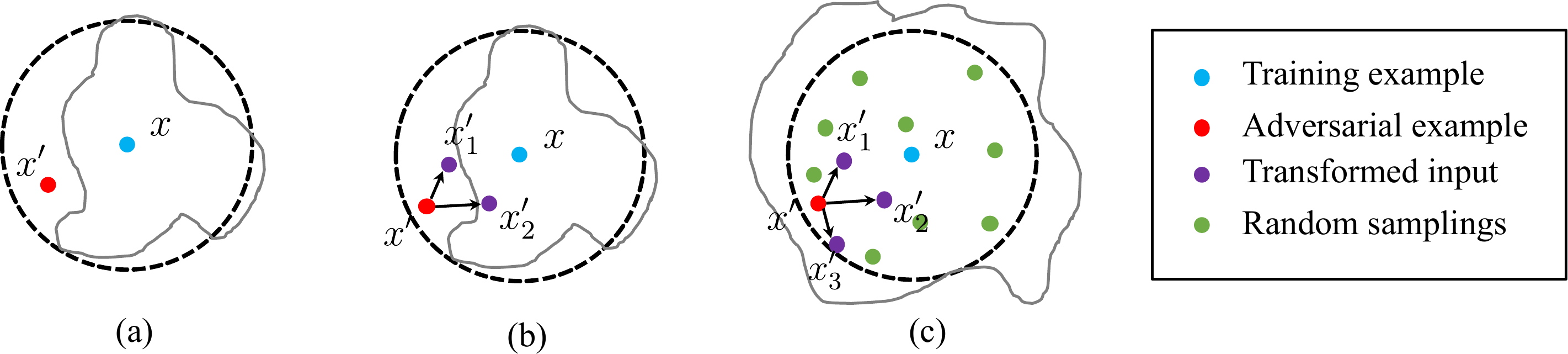}
	\caption{Decision boundary around normal example. }
	\label{fig:moti}
\end{figure}

There are many possible reasons why DNNs have vulnerabilities to adversarial attacks.
One of the pivotal factors comes from the internal robustness of neural networks. 
Given a normal example $x$, suppose $x$ is within the decision boundary in which the classifier can make a correct prediction, as seen in Fig.\ref{fig:moti}(a).
However, attackers can craft an adversarial example $x'$ in the neighborhood of $x$ such that the classifier will make a wrong prediction on $x'$. 

For word-level adversarial attacks, adversarial examples within neighborhood of normal examples are generally created by substituting parts of words in the text by their synonyms.
Therefore a judicious solution is to encode synonyms into the same embeddings\cite{wang2019natural}, and use the modified embeddings to train neural networks as well as testing.
However, this method does not enlarge the decision boundary too much because the number of training data is limited.
Thus under this encoding, a carefully crafted adversarial example may or may not go through the decision boundary.
From Fig.\ref{fig:moti}(b) we can see that under such encoding, adversarial example $x'$ may be mapped to $x_1'$(defense fail) or $x_2'$(defense success).

In this paper, we apply a different way to generate more robust word embeddings.
We randomly involve neighborhood examples of all training data into the model training process.
The neighborhood examples come from random synonym substitutions and they share the same label as the original example.
Thus the decision boundary of one example may be expanded to cover most unseen neighborhood examples including adversarial examples, as shown in Fig.\ref{fig:moti}(c). 
Note that we did not generate a large number of neighborhood examples for a training data because of the expensive training time.

To address this challenge, we adopt a dynamic synonym substitution strategy in the training process and the number of training data remains unchanged.
As presented in Fig.\ref{fig:moti}(c), a neighborhood example(a green circle) replaces the original example(the blue circle) to involve in the training process in an epoch. 
Thus different neighborhood examples are generated and work in different epochs.
In the test process, testing examples are also required to randomly substitute by their synonyms.
As a result, no matter an unseen example(may be adversarial) $x'$ is mapped to $x_1'$, $x_2'$ or $x_3'$, the model can also give the correct prediction.
We give the details of our framework in the next subsection.

\subsection{Framework Specification}
\label{sec-spec}

Given a set of training examples $\{x_i,y_i\}_N$, a text classification model $\mathcal{M}$ with parameter $\theta$, the objective of M is to minimize the negative log-likelihood:
\begin{equation}\label{equ:obj}
\min_\theta \left\{\mathcal{L}(\theta):=-\sum_{i}^{N}\text{log}\ P(y_i|x_i;\theta)\right\}
\end{equation}

To make the decision boundary more refined and ease the training load,
our approach do not generate many labeled examples in advance. 
We dynamically generate neighborhood examples instead of original examples in every epoch of the training process. 
To this end, we proposed a dynamic random synonym substitution based framework RSE that introduces a Random Substitution Encoder between the input and the embedding layer.
Then the training objective is to minimize:
\begin{equation}\label{equ:ourobj}
\min_\theta \left\{- \sum_{i}^{N} \rand_{\|\delta_i\|<\epsilon} \text{log}\ P(y_i| x_i+\delta_i;\theta)\right\}
\end{equation}
where $\rand$ denotes the random synonym substitution operation, and $\|\delta_i\|<\epsilon$ guarantees that the generated example $x_i+\delta_i$ stays in the neighborhood of $x_i$.

\begin{figure}[tb]
	\centering
	\setlength{\belowcaptionskip}{-0.5cm}
	\fontsize{9pt}{9pt}\selectfont
	\def\svgscale{0.78}
	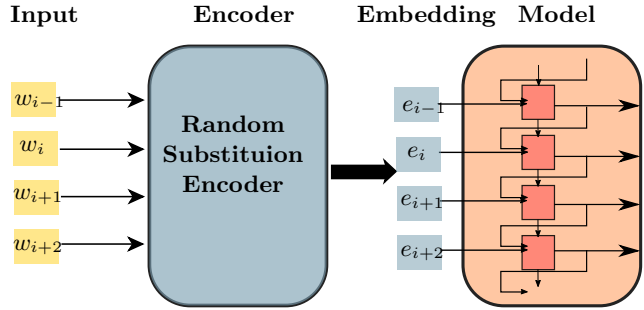
	\caption{The proposed RSE framework}
	\label{fig:frame}
\end{figure}

Since $\rand$ operation does not need an optimization, we can fuse random synonym substitution and encodes the inputs into a new embeddings in real time in the training process of model $\mathcal{M}$.
Fig.\ref{fig:frame} illustrates the representation of the proposed framework.

From Fig.\ref{fig:frame}, we can see that RSE reads an input text and encodes it to an embedding using a random synonym substitution algorithm.
For example, given an original example $x=w_1\cdots w_n$, RSE outputs a neighborhood $x'$ and feeds the embedding $x_e'=e_1\cdots e_n$ into the subsequent model $\mathcal{M}$.
Then model $\mathcal{M}$ is trained on the perturbed examples.
Here $\mathcal{M}$ can be one of any specific DNN models in NLP tasks and in this paper we focused on text classification model such as CNN and LSTM.

\subsection{Random Synonym Substitution Algorithm}
\label{sec-algo}
Next we introduce the details of RSE and the training process under the proposed framework.
In practice, to satisfy the constraints $\|\delta_i\|<\epsilon$, we adopt a substitution rate $sr$ instead of neighborhood radius $\epsilon$.
There are three steps to generate a neighborhood $x'$ for original example $x$. 
Firstly, we select a substitution rate $sr$ between a minimal rate $r_{min}$ and a maximal rate $r_{max}$.
Then we randomly sample a candidate words set $\mathcal{C}$ in which will be substituted.
Finally we randomly chose synonyms for all words in $\mathcal{C}$.
Algorithm\ref{alg:train} presents the details of these steps as well as the training process.

In the test stage, a test example also needs to be encoded under the proposed RSE in order to mitigate the possible adversarial noise.
For example, given a test example $x$, we firstly transform it to a representation of its neighborhood $x'$ by performing an algorithm(Lines 4-6 in Algorithm\ref{alg:train}).
Then the embedding is fed into the well-trained model to give a prediction.

\vspace{-0.5cm}
\begin{algorithm} 
	\caption{Training for the RSE framework}
	\label{alg:train}
	\KwIn{Training data $D=\{x_i,y_i\}_N$, model prameter $\theta$, minimal rate $r_{min}$ and maximal rate $r_{max}$}
	\For {epoch = $1\cdots N_{ep}$}{
		\For {minibatch $B\subset D$ }{
			\For {original example $x \in B$ }{
				Sample a substitution rate $sr$ between $r_{min}$ and $r_{max}$;\\
				Randomly sample candidate words set $\mathcal{C}$ \\
				Sample synonyms for all words in $\mathcal{C}$ to generate $x'$;\\
				Replace $x$ with $x'$ in $B$.\\
			}
			Update $\theta$ using gradient ascent of log-likelihood \ref{equ:ourobj} on \textit{minibatch} $B$.\\
		}
	}
\end{algorithm}
\vspace{-0.5cm}
%

\section{Experiments}
\label{sec-expe}
We evaluate the performance of the proposed RSE framework experimentally in this section.
We firstly present the experiment setup,
and then report the experiment results on three real-world datasets.
The results show that models under our RSE framework could achieve much better performance in defending adversarial examples.

\subsection{Experiment Setup}
\label{sec-setp}
In this subsection we give an overview of the datasets, target models, attack models and baselines used in our experiments.

\noindent \textbf{Datasets.} We test our RSE framework on three benchmark datasets: \emph{IMDB}, \emph{AG's News} and \emph{Yahoo! Answers}.

	\emph{IMDB}\cite{acl/MaasDPHNP11} is a dataset for binary sentiment classification  containing 25,000 highly polarized movie reviews for training and 25,000 for testing.
	
	\emph{AG's News}\cite{nips/ZhangZL15}  is extracted from news articles using only the title and description fields. 
	It contains 4 classes, and each class includes 30,000 training samples and 1900 testing examples.

	\emph{Yahoo! Answers}\cite{nips/ZhangZL15} is a topic classification dataset with 10 classes, which contains 4,483,032 questions and corresponding answers. 
	We sampled 150,000 training data and 5,000 testing data from the original 1,400,000 training data and 60,000 testing data for the following experiments.
	Each class contains 15,000 training data and 500 testing data respectively.

We also used padding length when preprocessing the input text. 
The padding length is decided by each datasets' average sentence length.
Table \ref{tab:1} lists the detailed description of the aforementioned datasets.

\vspace{-0.5cm}
\begin{table}[h]
	\footnotesize
	\caption{The statistic and preprocessing settings for each dataset }\label{tab:1}
	\begin{tabular}{|l|l|l|l|l|}
		\hline
		Dataset & 	\tabincell{c}{\# of \\Training samples} & \tabincell{c}{\# of\\ Testing samples} & \tabincell{c}{\# of \\Vocab words} & Padding length \\
		\hline
		IMDB & 25,000 & 25,000 & 80,000 & 300 \\
		AG's News & 120,000 & 7,600 & 80,000 & 50 \\
		Yahoo! Answers & 150,000 & 5,000 & 80,000 & 100 \\
		\hline
	\end{tabular}
\end{table}

\vspace{-0.5cm}
\noindent \textbf{Base Models.}
We used three main classic deep neural networks as base models in our RSE framework for text classification task: \emph{LSTM}, \emph{Bi-LSTM} and \emph{Word-CNN}.

\emph{LSTM} has a 100-dimension embedding layer, two LSTM layers where each LSTM cell has 100 hidden units and a fully-connected layer.

\emph{Bi-LSTM} also has a 100-dimension embedding layer, two bi-directional LSTM layers and a fully-connected layer. Each LSTM cell has 100 hidden units.

\emph{Word-CNN}\cite{emnlp/Kim14} has two embedding layers, one is static for pretrained word vectors, and another is non-static for training, three convolutional layers with filter size of 3, 4, and 5 respectively, one 1D-max-pooling layer and a fully-connected layer.

\noindent \textbf{Attack Models.}
We adopt three synonym substitution adversarial attack models to evaluate the effectiveness of defense methods.
We suppose attackers can obtain all testing examples of three datasets (IMDB, AG's News, Yahoo! Answers) and can call prediction interfaces of any models at any time.

\emph{Random}. We first randomly choose a set of candidate words that has synonyms. Then keep replacing the original word in the candidate set with a randomly synonym until the target model predicts wrong.

\emph{Textfool}\cite{kulynych2018evading} uses sorted word candidates based on the word similarity rank to replace with the synonym and keep perform until the target model predicts wrong. We will not use the typos substitution because we pay most attention to synonyms replacements attacks.

\emph{PWWS}\cite{RenDW18} is a greedy synonym
replacement algorithm called Probability Weighted
Word Saliency (PWWS) that considers the word
saliency as well as the classification probability. As the same, we only use synonym replacements but not specific named entities replacements.

\noindent \textbf{Baseline.}
We take \emph{NT}, \emph{AT} and \emph{SEM} as three baselines.
\emph{NT} is a normal training framework without taking any defense methods.
\emph{AT}\cite{corr/GoodfellowSS14} is an adversarial training framework, where extra adversarial examples are generated to train a robust model. 
We adopt the same adversarial training configurations as in \cite{wang2019natural}, which uses \emph{PWWS} to generate 10\% adversarial examples from each dataset for every normal trained neural network. 
Then the adversarial examples and original training examples are mixed for the training process.
\emph{SEM}\cite{wang2019natural} is an adversarial defense framework which inserts a fixed synonym substitution encoder before the input layer of the model.
We evaluate our framework and baseline frameworks by using \emph{LSTM}, \emph{Bi-LSTM} and \emph{Word-CNN} as base models respectively.

\subsection{Evaluations}
We evenly sampled each class from the origin test data to form 1,000 clean testing examples for every datasets.
Then these examples are used to generate adversarial examples by the above attack models, which will take \emph{NT}, \emph{AT}, \emph{SEM} and the proposed \emph{RSE} as the victim targets.

\vspace{-0.5cm}
\begin{table}[htbp]
	\footnotesize
	\centering
	\caption{The evaluation results under different settings.}
	{
		{
			\begin{tabular}{ll|rrr|rrr|rrr}
				\toprule
				\multirow{2}[4]{*}{Dataset} & \multirow{2}[4]{*}{Attack Model} & \multicolumn{3}{c}{LSTM (\%)} & \multicolumn{3}{c}{Bi-LSTM (\%)} & \multicolumn{3}{c}{Word-CNN (\%)} \\
				\cmidrule{3-11}          &       & \multicolumn{1}{c}{NT} & \multicolumn{1}{c}{AT} & \multicolumn{1}{c|}{RSE} & \multicolumn{1}{c}{NT} & \multicolumn{1}{c}{AT} & \multicolumn{1}{c|}{RSE} & \multicolumn{1}{c}{NT} & \multicolumn{1}{c}{AT} & \multicolumn{1}{c}{RSE} \\
				\midrule
				\multirow{4}[2]{*}{IMDB} & No attack   & \textbf{88.8} & 87.9  & 87.0  & \textbf{89.5} & 88.7  & 86.5  & 87.6  & 86.3  & \textbf{87.8} \\
				& Random & 80.6  & 77.8  & \textbf{83.1} & 81.5  & 79.2  & \textbf{81.9} & 77.5  & 75.0  & \textbf{83.0} \\
				& Textfool & 75.4  & 76.2  & \textbf{84.2} & 74.6  & 77.0  & \textbf{83.7} & 71.2  & 71.0  & \textbf{83.1} \\
				& PWWS  & 26.3  & 29.3  & \textbf{82.2} & 27.3  & 28.1  & \textbf{79.3} & 13.5  & 10.5  & \textbf{81.2} \\
				\midrule
				\multirow{4}[2]{*}{AG’s News} & No attack   & 90.5  & 92.6 & \textbf{92.9}  & \textbf{96.4} & 94.5  & 94.1  & 95.9  & \textbf{96.9} & 94.8 \\
				& Random & 84.7  & 87.9  & \textbf{89.2} & 91.4  & 89.7  & \textbf{92.2} & 91.6  & 92.6  & \textbf{93.1} \\
				& Textfool & 79.6  & 85.3  & \textbf{88.7} & 86.3  & 89.1  & \textbf{90.6} & 88.0  & \textbf{92.6} & 92.2 \\
				& PWWS  & 63.0  & 72.8  & \textbf{84.2} & 70.4  & 78.0  & \textbf{88.3} & 67.5  & 77.1  & \textbf{89.9} \\
				\midrule
				\multicolumn{1}{l}{\multirow{4}[2]{*}{\tabincell{l}{Yahoo! \\Answers }}} & No attack   & 72.5  & \textbf{73.1} & 72.1  & \textbf{73.2} & 72.7  & 71.8  & \textbf{71.2} & 66.0  & 70.1 \\
				& Random & 60.5  & 65.1  & \textbf{68.6} & 61.2  & 64.2  & \textbf{68.9} & 58.9  & 54.3  & \textbf{67.3} \\
				& Textfool & 58.9  & 63.4  & \textbf{67.4} & 60.4  & 63.6  & \textbf{67.1} & 57.9  & 56.2  & \textbf{66.4} \\
				& PWWS  & 29.1  & 39.3  & \textbf{64.3} & 26.3  & 39.2  & \textbf{64.6} & 28.8  & 26.8  & \textbf{62.6} \\
				\bottomrule
			\end{tabular}
	}}
	\label{tab:2}
\end{table}
\vspace{-0.5cm}

The key metrics to evaluate the performance of different defense frameworks in this paper are \emph{Accuracy}, \emph{Accuracy Shift} and \emph{Attack-Success Rate}.
\emph{Accuracy} refers to the ratio that the number of correctly predicted examples against the total number of testing examples.
\emph{Accuracy Shift} refers to the reduced accuracy before and after the attack.
\emph{Attack-Success Rate} is defined by the number of successfully attacked examples by attack models against the number of correctly predicted examples with no attack.
It can be computed by $(Accuracy\ Shift) /( No\ attack\ Accuracy)$.
The better defense performance the target model has, the lower Attack-Success Rate the attacker gets. 

Table \ref{tab:2} shows the accuracy results of base models(\emph{LSTM}, \emph{Bi-LSTM} and \emph{Word-CNN}) against various attack models (\emph{Random}, \emph{Textfool}, \emph{PWWS}) under \emph{NT}, \emph{AT}, and the proposed \emph{RSE} defense framework.
For each base model with each dataset, we highlight the highest classification accuracy for different defense frameworks in \textbf{bold} to indicate the best defense performance.

From table \ref{tab:2}, we can see the following observations when looking at each box to find the best accuracy result:
\begin{enumerate}
	\item When there is no attack, either NT or AT usually has the best accuracy. But under other attack models, our RSE framework can get the best accuracy.
	\item For each column in each box, target models have the lowest accuracy under PWWS attack, which demonstrates PWWS is the most effective attack model.
	The accuracy of NT and AT drop significantly under PWWS attack.
	But RSE has the best defense performance since the accuracy loss is very small compared with `No attack'.
	\item Under different settings (various datasets, attack models and base models), our RSE framework has a better performance with few accuracy decrease. 
	This demonstrates the generalization of RSE framework to strengthen a robust deep neural network against synonym adversarial attacks.
\end{enumerate}



\vspace{-0.8cm}
\begin{table}[htbp]
	\footnotesize
	\centering
	\caption{
		SEM VS. RSE under PWWS attack model.
	} 
	\begin{tabular}{clcc|cc|cc}
		\toprule
		\multirow{2}[4]{*}{Metric \%} & \multirow{2}[4]{*}{Base Model} & \multicolumn{2}{c}{IMDB} & \multicolumn{2}{c}{AG’s News} & \multicolumn{2}{c}{Yahoo! Answers } \\
		\cmidrule{3-8}          &       & SEM   & RSE   & SEM   & RSE   & SEM   & RSE \\
		\midrule
		\multirow{3}[2]{*}{\tabincell{c}{Before-Attack \\Accuracy}} & LSTM  & 86.8  & 87.0  & 90.9  & 92.9  & 69.0  & 72.1 \\
		& Bi-LSTM & 87.6  & 86.5  & 90.1  & 94.1  & 70.2  & 71.8 \\
		& Word-CNN & 86.8  & 87.8  & 88.7  & 94.8  & 65.8  & 70.1 \\
		\midrule
		\multirow{3}[2]{*}{\tabincell{c}{After-Attack \\Accuracy}} & LSTM  & 77.3  & 82.2  & 85.0  & 84.2  & 54.9  & 64.3 \\
		& Bi-LSTM & 76.1  & 79.3  & 81.1  & 88.3  & 57.2  & 64.6 \\
		& Word-CNN & 71.1  & 81.2  & 67.6  & 89.9  & 52.6  & 62.6 \\
		\midrule
		\multirow{3}[2]{*}{Accuracy Shift} & LSTM  & 9.5   & \textbf{4.8 } & \textbf{5.9 } & 8.7   & 14.1  & \textbf{7.8 } \\
		& Bi-LSTM & 11.5  & \textbf{7.2 } & 9.0   & \textbf{5.8 } & 13.0  & \textbf{7.2 } \\
		& Word-CNN & 15.7  & \textbf{6.6 } & 21.1  & \textbf{4.9 } & 13.2  & \textbf{7.5 } \\
		\midrule
		\multirow{3}[2]{*}{\tabincell{c}{Attack-Success \\Rate}} & LSTM  & 10.94 & \textbf{5.52} & \textbf{6.49} & 9.36  & 20.43 & \textbf{10.82} \\
		& Bi-LSTM & 13.13 & \textbf{8.32} & 9.99  & \textbf{6.16} & 18.52 & \textbf{10.03} \\
		& Word-CNN & 18.09 & \textbf{7.52} & 23.79 & \textbf{5.17} & 20.06 & \textbf{10.70} \\
		\bottomrule
	\end{tabular}%
	\label{tab:3}%
\end{table}%
\vspace{-0.5cm}

\noindent\textbf{RSE vs. SEM}.
We also compared SEM with our RSE as shown in table \ref{tab:3}. 
Please note that we evaluate the performance of RSE and SEM only under PWWS attack model on three datasets because: 
(1) PWWS has the strongest attacking efficacy; and 
(2) SEM has no opened source codes yet and we directly cite the results in \cite{wang2019natural} under the same experimental settings. 

It can be seen from Table \ref{tab:3} that the average After-Attack Accuracy of RSE is higher than SEM for about 5\%-10\%. 
We also compared the Accuracy Shift of SEM and RSE since the parameters of each base model may be different.
We find out that except for AG's News dataset with LSTM model, models under RSE have smaller Accuracy Shift, and the shifts are stable with only 5\% decrease in average.
But for SEM the decrease is about 10\% in average. 
For AG's News dataset with Word-CNN model, the Accuracy Shift reaches 21.1\% and thus the performance of the model is unacceptable.
We can also see that our RSE has lower Attack-Success Rate for nearly all settings.
This means that it is more difficult for PWWS attacker to craft adversarial examples under RSE framework.

\noindent\textbf{Substitution Rate}.
When crafting an adversarial example, it is better to add smaller perturbations. 
Thus noisy rate is an important metric in adversarial attack.
It means that the crafted examples may not be imperceptible if the noisy rate is high.
On the contrary, the defense mechanism is better if it causes the attacker have to add more noise to success. 
Thus in this paper we introduce Substitution Rate as a metric, which is defined as the number of substituted words against the sentence length.
The better performance the defense framework has, the more the substituted words the attack model costs. 

\vspace{-0.5cm}
\begin{table}[htbp]
	\footnotesize
	\centering
	\caption{
		Performance on Substitution Rate.
	}
	\begin{tabular}{llrrr|rrr|rrr}
		\toprule
		\multicolumn{1}{l}{\multirow{2}[4]{*}{Dataset}} & \multirow{2}[4]{*}{Attacker} & \multicolumn{3}{c}{LSTM (\%)} & \multicolumn{3}{c}{Bi-LSTM (\%) } & \multicolumn{3}{c}{Word-CNN (\%)} \\
		\cmidrule{3-11}          &       & \multicolumn{1}{l}{NT} & \multicolumn{1}{l}{AT} & \multicolumn{1}{l|}{RSE} & \multicolumn{1}{l}{NT} & \multicolumn{1}{l}{AT} & \multicolumn{1}{l|}{RSE} & \multicolumn{1}{l}{NT} & \multicolumn{1}{l}{AT} & \multicolumn{1}{l}{RSE} \\
		\midrule
		\multirow{2}[2]{*}{IMDB} & Textfool & 17.98 & 18.22 & \textbf{20.09} & 17.63 & 18.26 & \textbf{20.06} & 17.57 & 17.35 & \textbf{20.03} \\
		& PWWS  & 10.54 & 11.23 & \textbf{19.13} & 10.55 & 11.25 & \textbf{18.12} & 6.41  & 5.36  & \textbf{17.99} \\
		\midrule
		\multirow{2}[2]{*}{AG’s News} & Textfool & 20.93 & 21.27 & \textbf{22.18} & 21.18 & 21.39 & \textbf{22.11} & 22.09 & 21.59 & \textbf{22.17} \\
		& PWWS  & 19.07 & 20.88 & \textbf{21.66} & 20.09 & 21.44 & \textbf{22.20} & 19.35 & 20.75 & \textbf{23.08} \\
		\midrule
		\multirow{2}[2]{*}{\tabincell{l}{Yahoo! \\Answers}} & Textfool & 13.69 & \textbf{14.19} & 13.25 & 13.59 & \textbf{14.29} & 13.38 & 13.88 & \textbf{13.71} & 13.50 \\
		& PWWS  & 10.28 & 12.47 & \textbf{16.09} & 9.79  & 12.85 & \textbf{16.30} & 10.30 & 9.74  & \textbf{16.21} \\
		\bottomrule
	\end{tabular}%
	\label{tab:4}%
\end{table}%
\vspace{-0.5cm}

The table \ref{tab:4} shows the Substitution Rate of each base model without (NT) or with defend frameworks (AT and RSE). 
We could not list the results of SEM since they did not report in \cite{wang2019natural}.
From table \ref{tab:4} it could be seen that the Substitution Rate for attacking the models with RSE is over 20\% in most cases, better than NT and AT.
So we can safely conclude that RSE makes the attackers pay more cost for perturbing origin sentences. 

\section{Conclusion and Future Work}
\label{sec-cons}
In this paper, we propose a defense framework called RSE to protect text classification models against word-level adversarial attacks.
With this framework,
a random synonym substitution encoder is fused into the deep neural network to endow base models with robustness to adversarial examples.
And a corresponding training algorithm is also proposed.
Extensive experiments on three popular real-world datasets demonstrate the effectiveness of our framework on defense of word-level adversarial attacks.
In the future, we will explore how the parameters of our algorithm impact the performance and transfer our RSE framework into other typical NLP tasks, e.g., Machine Translation and Question \& Answer, to protect deep models from word-leval adversarial attacks.

\section*{Acknowledgments}
This work was supported by the National Natural Science Foundation of China (Grant No.61802124).

%

\bibliographystyle{plain}
\bibliography{ref}

\begin{thebibliography}{10}

\bibitem{AlzantotSEHSC18}
Moustafa Alzantot, Yash Sharma, Ahmed Elgohary, Bo{-}Jhang Ho, Mani~B.
  Srivastava, and Kai{-}Wei Chang.
\newblock Generating natural language adversarial examples.
\newblock In {\em Proceedings of the 2018 Conference on Empirical Methods in
  Natural Language Processing}, pages 2890--2896, 2018.

\bibitem{EbrahimiLD18}
Javid Ebrahimi, Daniel Lowd, and Dejing Dou.
\newblock On adversarial examples for character-level neural machine
  translation.
\newblock In {\em Proceedings of the 27th International Conference on
  Computational Linguistics}, pages 653--663, 2018.

\bibitem{EbrahimiRLD18-2}
Javid Ebrahimi, Anyi Rao, Daniel Lowd, and Dejing Dou.
\newblock Hotflip: White-box adversarial examples for text classification.
\newblock In {\em Proceedigs of the 56th Annual Meeting of the Association for
  Computational Linguistics, {ACL} 2018}, pages 31--36, 2018.

\bibitem{GaoLSQ18}
Ji~Gao, Jack Lanchantin, Mary~Lou Soffa, and Yanjun Qi.
\newblock Black-box generation of adversarial text sequences to evade deep
  learning classifiers.
\newblock In {\em 2018 {IEEE} Security and Privacy Workshops, {SP} Workshops
  2018}, pages 50--56.

\bibitem{corr/GoodfellowSS14}
Ian~J. Goodfellow, Jonathon Shlens, and Christian Szegedy.
\newblock Explaining and harnessing adversarial examples.
\newblock In {\em 3rd International Conference on Learning Representations,
  {ICLR} 2015}, 2015.

\bibitem{emnlp/Kim14}
Yoon Kim.
\newblock Convolutional neural networks for sentence classification.
\newblock In {\em Proceedings of the 2014 Conference on Empirical Methods in
  Natural Language Processing, {EMNLP} 2014}, pages 1746--1751, 2014.

\bibitem{kuleshov2018adversarial}
Volodymyr Kuleshov, Shantanu Thakoor, Tingfung Lau, and Stefano Ermon.
\newblock Adversarial examples for natural language classification problems.
\newblock 2018.

\bibitem{kulynych2018evading}
Bogdan Kulynych, Jamie Hayes, Nikita Samarin, and Carmela Troncoso.
\newblock Evading classifiers in discrete domains with provable optimality
  guarantees.
\newblock {\em arXiv preprint arXiv:1810.10939}, 2018.

\bibitem{0002LSBLS18}
Bin Liang, Hongcheng Li, Miaoqiang Su, Pan Bian, Xirong Li, and Wenchang Shi.
\newblock Deep text classification can be fooled.
\newblock In {\em Proceedings of the Twenty-Seventh International Joint
  Conference on Artificial Intelligence}, pages 4208--4215, 2018.

\bibitem{acl/MaasDPHNP11}
Andrew~L. Maas, Raymond~E. Daly, Peter~T. Pham, Dan Huang, Andrew~Y. Ng, and
  Christopher Potts.
\newblock Learning word vectors for sentiment analysis.
\newblock In {\em The 49th Annual Meeting of the Association for Computational
  Linguistics: Human Language Technologies}, pages 142--150, 2011.

\bibitem{sp/PapernotM0JS16}
Nicolas Papernot, Patrick~D. McDaniel, Xi~Wu, Somesh Jha, and Ananthram Swami.
\newblock Distillation as a defense to adversarial perturbations against deep
  neural networks.
\newblock In {\em {IEEE} Symposium on Security and Privacy}, pages 582--597,
  2016.

\bibitem{PruthiDL19}
Danish Pruthi, Bhuwan Dhingra, and Zachary~C. Lipton.
\newblock Combating adversarial misspellings with robust word recognition.
\newblock In {\em Proceedings of the 57th Conference of the Association for
  Computational Linguistics, {ACL} 2019}, pages 5582--5591.

\bibitem{acl/PruthiDL19}
Danish Pruthi, Bhuwan Dhingra, and Zachary~C. Lipton.
\newblock Combating adversarial misspellings with robust word recognition.
\newblock In {\em Proceedings of the 57th Conference of the Association for
  Computational Linguistics, {ACL} 2019}, pages 5582--5591, 2019.

\bibitem{RenDHC19}
Shuhuai Ren, Yihe Deng, Kun He, and Wanxiang Che.
\newblock Generating natural language adversarial examples through probability
  weighted word saliency.
\newblock In {\em Proceedings of the 57th Conference of the Association for
  Computational Linguistics, {ACL} 2019}, pages 1085--1097, 2019.

\bibitem{RenDW18}
Yuanhang Ren, Ye~Du, and Di~Wang.
\newblock Tackling adversarial examples in {QA} via answer sentence selection.
\newblock In {\em Proceedings of the Workshop on Machine Reading for Question
  Answering@ACL 2018}, pages 31--36, 2018.

\bibitem{wang2019natural}
Xiaosen Wang, Hao Jin, and Kun He.
\newblock Natural language adversarial attacks and defenses in word level.
\newblock {\em arXiv preprint arXiv:1909.06723}, 2019.

\bibitem{nips/ZhangZL15}
Xiang Zhang, Junbo~Jake Zhao, and Yann LeCun.
\newblock Character-level convolutional networks for text classification.
\newblock In {\em Annual Conference on Neural Information Processing Systems
  2015}, pages 649--657, 2015.

\end{thebibliography}

\end{document}